\def\year{2020}\relax
\documentclass[letterpaper]{article} % DO NOT CHANGE THIS
\usepackage{aaai20}  % DO NOT CHANGE THIS
\usepackage{times}  % DO NOT CHANGE THIS
\usepackage{helvet} % DO NOT CHANGE THIS
\usepackage{courier}  % DO NOT CHANGE THIS
\usepackage[hyphens]{url}  % DO NOT CHANGE THIS
\usepackage{graphicx} % DO NOT CHANGE THIS
\usepackage{graphicx}  % DO NOT CHANGE THIS
\usepackage{xspace}
\title{Towards Intelligent Robotic Process Automation for BPMers}

\author{Simone Agostinelli, Andrea Marrella, Massimo Mecella\\
Sapienza Universit\`{a} di Roma\\ %If you have multiple authors and multiple affiliations
\{agostinelli,marrella,mecella\}@diag.uniroma1.it % email address must be in roman text type, not monospace or sans serif
}

\usepackage{amsfonts}
\usepackage[title]{appendix}
\usepackage{tikz}
\usepackage{pgfplots}
\usepackage{pgfplotstable}
\usepackage{booktabs}
\usepackage[defblank]{paralist}
\usepackage{enumitem}
\usepackage{verbatim}
\usepackage{amssymb}
\usepackage{rotating}
\usepackage{url}
\usepackage{tabularx}
\usepackage{booktabs}
\usepackage{amsmath}

\usepackage{color}
\usepackage{verbatim}
\usepackage{epsfig}
\usepackage{amssymb}
\usepackage{amsmath}
\usepackage{rotating}
\usepackage{enumitem}
\usepackage{url}
\usepackage{tabularx}
\usepackage{booktabs}
\usepackage{fixltx2e}
\usepackage[clock,electronic]{ifsym}
\usepackage{eurosym}
\usepackage{subfig}
\usepackage{xspace}
\usepackage[defblank]{paralist}
\usepackage{fixltx2e}
\usepackage{multirow}
\usepackage{booktabs}
\usepackage{amssymb}  % additional symbols (there are more packages)
\usepackage{gensymb}
\usepackage{amsmath}
\usepackage{makecell}
\usepackage[ruled, vlined, linesnumbered]{algorithm2e}
\newcommand{\removelatexerror}{\let\@latex@error\@gobble}
\usepackage{wrapfig}
\usepackage{adjustbox}
\usepackage{dsfont}

\begin{document}

\sloppypar

\maketitle

\begin{abstract}
Robotic Process Automation (RPA) is a fast-emerging automation technology that sits between the fields of Business Process Management (BPM) and Artificial Intelligence (AI), and allows organizations to automate high volume routines. RPA tools are able to capture the execution of such routines previously performed by a human users on the interface of a computer system, and then emulate their enactment in place of the user by means of a software robot.
Nowadays, in the BPM domain, only simple, predictable business processes involving routine work can be automated by RPA tools in situations where there is no room for interpretation, while more sophisticated work is still left to human experts.
In this paper, starting from an in-depth experimentation of the RPA tools available on the market, we provide a classification framework to categorize them on the basis of some key dimensions. Then, based on this analysis, we derive four research challenges and discuss prospective approaches necessary to inject intelligence into current RPA technology, in order to achieve more widespread adoption of RPA in the BPM domain.
%\Keywords{Robotic Process Automation \and Business Process Management \and Process Mining \and Human Computer Interaction \and Automated Planning}
\end{abstract}

\section{Introduction}
\label{sec:intro}

The recent developments in Artificial Intelligence (AI) force us to continuously revisit the debate on \emph{what should be automated and what should be done by humans}.
One of these developments is Robotic Process Automation (RPA), a fast-emerging automation approach that use \emph{software robots} (or simply \emph{SW robots}) to mimic and replicate the execution of highly repetitive tasks performed by humans in their application's user interface (UI).
SW robots are mainly used for automating office tasks in operations like accounting, billing and customer service.
Typical tasks are: extract semi-structured data from documents, read and write from/to databases, copy and paste data across cells of a spreadsheet, open e-mails and attachments, fill in forms, make calculations, etc. \cite{willcocks2015function,willcocks2016service}.

Despite the capabilities of SW robots, the RPA technology is still considered to be in its infancy \cite{vanderAalst2018}, even if similar solutions have been around for a long time. For instance, since the mid-nineties, closely related to SW robots, \emph{chatbots} have been used for years to accept voice-based or keyboard inputs and guide customers to find relevant information in web-based applications \cite{hill2015real}.
Similarly, in the same years, there has been some interest in the financial industry around a preliminary form of RPA, called Straight Through Processing (STP) \cite{van2004workflow}. The objective of STP was to speed up financial transactions allowing to automate several repetitive back-office functions, thus reducing the manual process for employees to enter the same information repeatedly during the processing of a transaction.
Differently from STP and chatbots, RPA can be seen as an evolution of traditional \textit{screen scraping} solutions \cite{bisbal1999legacy}, which sought to visualize screen display data from legacy applications (having no means for automated interfacing) in order to display such data using modern UIs. The strength of RPA is that it does not replace existing applications or manipulate their code, but rather works with them similarly to a human user.

In recent years, there has been an increased interest around RPA in the Business Process Management (BPM) domain \cite{kirchmer2017robotic}. BPM is an active area of research based on the observation that each product and/or service that an organization offers is the outcome of a number of performed activities. Business processes (BPs) are the key instrument for organizing such activities and improving the understanding of their interrelationships. Nowadays, BPs are enacted in many complex industrial (e.g., manufacturing, logistics, retail) and non-industrial (e.g., emergency management, healthcare, smart environments) domains through a dedicated generation of information systems, called Process Management Systems (PMSs) \cite{ReichertBook2012}. However, while conducting a BPM project is often considered too expensive because its ``top-down'' approach that forces to develop the PMS from scratch (and system integration is expensive), the promise of RPA is to rely on an approach where, instead of redesigning existing information systems (that remain unchanged), humans are replaced by SW robots in the execution of those BPs involving routine work. This allows knowledge workers to have more time for value added tasks.  
%
%that emerges in the last five years as a set of software tools and platforms that can automate tasks on rules-based business process.
%
In the research literature, a number of case studies have shown that RPA technology can concretely lead to improvements in efficiency for BPs involving routine work in large companies, such as O2 and Vodafone \cite{lacity2015robotic,Aguirre2017,geyer2018process}.

Despite this growing attention around RPA, when considering the state-of-the-art technology, it becomes apparent that the current generation of RPA tools is driven by predefined rules and manual configurations made by expert users rather than by AI \cite{aitiptoes}. According to \cite{vanderAalst2018}, to achieve a more widespread adoption in the BPM domain, RPA needs to become ``smarter''. In a nutshell, with the use of AI techniques, more complex and less defined BPs could be fully supported by the RPA technology.
For example, by observing human problem resolving unexpected system behavior (e.g., in case of system errors, changing forms, etc.), RPA tools can adapt and learn how to handle non-standard cases.
After all, the Gartner Hype Cycle for AI published in 2018\footnote{\url{https://www.gartner.com/en/documents/3883863-hype-cycle-for-artificial-intelligence-2018}} places RPA as one of the technologies at the peak of the hype cycle, meaning that there are nowadays deep expectations on what RPA will be able to deliver to the AI community.

%Second, RPA aims to be robust with respect to changes ofthe underlying information systems. When the layout of an electronic form changes, but the key content remains unchanged, the RPA software should adapt just as humans do.

%research aims at tackling this issue by realizing intelligent methods and algorithms to the automated identification of softbot behaviours from RPA logs and their intelligent orchestration as complex workflows, without the need of manual configurations.

Starting from the above discussion, in this paper we first identify and test ten RPA tools available on the market and categorize them by means of a classification framework. The results of the classification allow us to derive four research challenges and identify prospective approaches required to evolve RPA towards AI in the context of BPM.

\section{Background}
\label{background}

%In the mid-nineties, there was some interest in the financial industry around a preliminary form of RPA, called Straight Through Processing (STP) \cite{van2004workflow}. The term refers to specific workflows targeted to speed up financial transactions by processing that could be performed without any human involvement. The objective of STP was to allow companies to have the same information be streamlined through a transaction across multiple points, thus reducing the manual process for employees of having to keep entering the same information over and over again or checking to ensure a transaction fully processed, which raised chances for errors. Unfortunately, according to \cite{vanderAalst2018}, STP turned out to be applicable to only a few highly structured workflows, causing a rapid decrease of the hype around this approach.

The state-of-the-art in RPA is plenty of recent works that are focused on optimizing specific BPM aspects of a RPA project. The work of Bosco et al. \cite{RPM} focuses on the automated identification of candidate routines to robotize. The work of Gao et al. \cite{COOPIS2019} proposes a self-learning approach to automatically detect high-level RPA-rules, from captured historical low-level user behavior. The work of Jimenez-Ramirez and Reijers \cite{10.1007/978-3-030-21290-2_28} proposes a method for the early stages of a RPA project. The work of Leno \cite{lenoaction} presents a method to record the performed user actions in Excel and Chrome Web browser in a log, in order to enable process mining for RPA.
Finally, the works \cite{lacity2015robotic,Aguirre2017,geyer2018process} discuss the enactment of three different RPA case studies in large companies.

Excluding the works \cite{RPM} and \cite{COOPIS2019}, the majority of the above contributions focus on refining some existing features of a RPA project, while none of them is targeted to identify and tackle the concrete research challenges of RPA to enable its adoption in BPM on a large scale. To fill this gap, the first step of this contribution is to identify and test the real RPA tools available on the market categorizing them by means of a classification framework.

Most of the actual deployments of RPA are industry-specific, e.g., for financial and business services \cite{market}. According to \cite{RPA}, nowadays, the market of RPA solutions includes more than 50 vendors developing tools having different prices and features. Among them, we identified 10 vendors that offer to freely try their RPA tools, i.e., without the need to pay any license. The RPA tools in question are:

\begin{itemize}
\item \textit{Automation Anywhere}\footnote{\url{https://www.automationanywhere.com/}}
\item \textit{AssistEdge}\footnote{\url{https://www.edgeverve.com/assistedge/}}
\item \textit{G1ANT}\footnote{\url{https://g1ant.com/}}
\item \textit{Kryon}\footnote{\url{https://www.kryonsystems.com/}}
\item \textit{Rapise}\footnote{\url{https://www.inflectra.com/Rapise/}}
\item \textit{TagUI}\footnote{\url{https://github.com/kelaberetiv/TagUI}}
\item \textit{UiPath}\footnote{\url{https://www.uipath.com/}}
\item \textit{VisualCron}\footnote{\url{https://www.visualcron.com/}}
\item \textit{WinAutomation}\footnote{\url{https://www.winautomation.com/}}
\item \textit{WorkFusion}\footnote{\url{https://www.workfusion.com/}}
\end{itemize}

We analyzed each of the above tools leveraging a dedicated case study based on a Purchase-to-Pay process obtained from a SAP ERP system (the same one used in \cite{geyer2018process}), which includes many standardized and highly repetitive transactions with potential for automation. The data of the case study covers around 400.000 purchase orders created over one year, and can be accessed after registration on the Celonis Academic Cloud \cite{celonis}.

After selecting the target process to automate, we employed the selected tools to design and train various SW robots, by recording the manual steps of the process. This has allowed us to identify a list of common tasks that must be performed to conduct a RPA project:

\begin{enumerate}
\item Determine which process steps (also called \emph{routines}) are good candidates to be automated.
\item Model the selected routines in the form of \emph{flowchart diagrams}, which involve the specification of the actions, routing constructs (e.g., parallel and alternative branches), data flow, etc. that define the behavior of a SW robot.
\item Record the mouse/key events that happen on the UI of the user's computer system. This information is associated with the actions of a routine, enabling it to emulate the recorded human activities by means of a SW robot.
\item Develop each modeled routine by generating the software code required to concretely enact the associated SW robot on a target computer system.
\item Deploy the SW robots in their environment to perform their actions. According to \cite{10.1007/978-3-030-21290-2_28}, it should be noted that RPA is often characterized by the lacking of a testing environment; only the production environment is available.
\item Monitor the performance of SW robots to detect bottlenecks and exceptions.
\item Maintain the routines, which takes into account each SW robot's performance and error cases. The outcomes of this phase enable a new analysis and design cycle to enhance the SW robots \cite{10.1007/978-3-030-21290-2_28}.
\end{enumerate} 
\section{Classification Framework}
\label{classification-framework}

\begin{table*}[t!]
\caption{Results of the application of the classification framework}
\label{rpa-table}
\begin{adjustbox}{width=1.0\textwidth,center=\textwidth}
\newcolumntype{C}{>{\centering\arraybackslash}p{24mm}}
\pgfplotstabletypeset[
column type/.add={|}{|},% results in ’|c’
every column/.code={
\ifnum\pgfplotstablecol>20
\pgfkeysalso{column type/.add={}{|}}%
\fi},
every head row/.style={
before row={%
\toprule
\multicolumn{1}{|c||}{\textbf{Tool}} &
%\multicolumn{1}{|c|}{\textbf{Accessibility}} &
%\multicolumn{4}{c||}{\textbf{Supported OSs}} &
\multicolumn{2}{c||}{\textbf{SW Arch.}} &
\multicolumn{3}{c||}{\textbf{Coding}} &
\multicolumn{3}{c||}{\textbf{Recording}} &
\multicolumn{2}{c||}{\textbf{Self Learning}} &
\multicolumn{1}{c||}{\textbf{Autom.}} &
\multicolumn{2}{c||}{\textbf{Routine comp.}} &
\multicolumn{1}{c|}{\textbf{Log}} \\
},
after row=\midrule,
},
every last row/.style={
after row=\bottomrule},
columns/extra/.style ={column name=},
columns/authors/.style ={column name=},
columns/extra2/.style = {column name=\textbf{quality}},
columns/web/.style ={column name=Web},
columns/desktop/.style ={column name=Desktop},
%columns/mobile/.style ={column name=Mobile},
columns/others/.style ={column name=Others},
%columns/windows/.style ={column name=Windows},
%columns/linux/.style ={column name=Linux},
%columns/macos/.style ={column name=macOS},
columns/clientserver/.style ={column name=Cl.-Server},
columns/standalone/.style ={column name=St.-alone},
columns/strong/.style ={column name=Strong},
columns/low/.style ={column name=Low},
columns/gui/.style ={column name=GUI},
columns/attended/.style ={column name=\textbf{type}},
%columns/unattended/.style ={column name=Unatt.},
%columns/hybrid/.style ={column name=Hybrid},
columns/manual/.style ={column name=Manual},
columns/automated/.style ={column name=Autom.},
columns/intra/.style ={column name=Intra-rout.},
columns/inter/.style ={column name=Inter-rout.},
col sep=&,row sep=\\,
string type,
]{
%person &
extra  %& authors
%& windows & linux & macos & mobile
&
clientserver & standalone & strong & low & gui & web & desktop & others & intra & inter & attended %& unattended & hybrid
& manual & automated & extra2 \\
%%%%%%
%%%%%%
%%%%%%
%Argos Lab & free trial & & & \checkmark & \checkmark & \checkmark & \checkmark & & & & \checkmark & & \\
Aut. Anyw. %& community
& \checkmark &  &  & \checkmark & \checkmark & \checkmark & \checkmark & \checkmark & & & Hybrid & \checkmark & & $\bigstar \bigstar \bigstar$\\
AssistEdge %& free trial
&  & \checkmark &  & \checkmark & \checkmark &  &  & \checkmark & & & Hybrid & \checkmark & & $\bigstar$ \\ %or $\bigstar \bigstar$\\
G1ANT %& community
& & \checkmark & \checkmark & & & \checkmark & \checkmark & & & & Hybrid & \checkmark & & $\bigstar$\\ %or $\bigstar \bigstar$\\
Kryion %& free trial
& \checkmark &  & & \checkmark & \checkmark &  & \checkmark & \checkmark & & & Hybrid & \checkmark & & $\bigstar \bigstar$\\ %discovery of automatable routines marlon dumas\\
Rapise %& free trial
& \checkmark &  &  & \checkmark & \checkmark & \checkmark & \checkmark & & & & Hybrid & \checkmark & & $\bigstar$\\
TagUI %& community
& & \checkmark  & \checkmark & & & \checkmark & \checkmark & & & & Hybrid & \checkmark & & $\bigstar$\\
UiPath %& community
& \checkmark &  &  & \checkmark & \checkmark & \checkmark & \checkmark & \checkmark & & & Hybrid & \checkmark & & $\bigstar \bigstar$\\
VisualCron %& free trial
& \checkmark &  & & & \checkmark &  & & & & & Attended & \checkmark & & $\bigstar \bigstar \bigstar$\\ %$\bigstar \bigstar$ or \\
WinAutom. %& free trial
& \checkmark &  &  & \checkmark & \checkmark & \checkmark & \checkmark & & & & Hybrid & \checkmark& & $\bigstar \bigstar$\\ %or $\bigstar \bigstar \bigstar$\\
WorkFusion %& community
& \checkmark  &  &  & \checkmark & \checkmark & \checkmark & \checkmark &\checkmark & & & Hybrid & \checkmark & & $\bigstar \bigstar$\\ %$\bigstar$
%%
%%%%%%
}
\end{adjustbox}
\end{table*}

We tested the selected RPA tools with our case study performing the tasks to conduct a RPA project. This has allowed us to realize a \emph{classification framework} for RPA tools, which consists of the following key dimensions:
\begin{itemize}
%\item \textbf{Accessibility}: the way under which the tools can be tried: either by a \textit{free trial} or through a \textit{community edition}; %in che modo è possibile ottenere il tool, se tramite una prova gratuita o scaricando l'edizione community;
%
%\item \textbf{Supported Operating Systems (OSs)}: The OSs supported by any tool, e.g., \textit{Windows}, \textit{Linux}, \textit{MacOS}, or mobile OSs, e.g., \textit{Android} and \textit{iOS}.
%
\item \textbf{Software (SW) Architecture}: The specific SW architecture adopted by the tool: either \textit{Stand-alone} or \textit{Client-Server}.
\item \textbf{Coding features}: The behavior of SW robots can be classified as:
\begin{itemize}
\item \textit{Strong coding}: based on the realization of explicit programming scripts, often with the support of a command-line interface (CLI), which instructs the SW robots about the routines to emulate;
\item \textit{Graphical User Interfaces (GUIs)}: user friendly environments providing drag \& drop facilities to build the flowchart of the routines to emulate;
\item \textit{Low-code tools}: GUIs that -- in addition to drag \& drop facilities -- provide low-coding functionalities to semi-automatically create software code.
\end{itemize}
\item \textbf{Recording facilities}: The actions performed by a human within the software tool can be recorded with:
%
%the way through which the software robots learn the user actions on the UI of interest.
\begin{itemize}
\item \textit{Web recording}: detection of user actions performed on a web browser;
\item \textit{Desktop recording}: detection of user actions performed on a desktop UI;
\item \textit{Others}: some RPA tools do not support neither web nor desktop recording. Nonetheless, they offer recording tools that work on specific applications only, such as Excel, Acrobat, SAP and Citrix. Some RPA tools provide also traditional screen-scraping recording.
\end{itemize}
\item \textbf{Self Learning}: The ability of the RPA tool to automatically understand which user actions belong to which routines (\textit{Intra-routine} learning), and which routines are good candidates for the automation (\textit{Inter-routine} learning).
%\begin{itemize}
%\item \textit{Intra-routine}: the software robots need to understand which actions belong to which routine;
%\item \textit{Inter-routine}: discrimination of the routines to be emulated through software robots.
%\end{itemize}
%
\item \textbf{Automation type}: SW robots can either interact with users and/or act independently. This leads to three different categories of automation:
\begin{itemize}
\item \textit{Attended}: the SW robots constantly require interaction with the users;%It is also called Robotic Desktop Automation.;
\item \textit{Unattended}: the SW robots act like batch processes, i.e., manual intervention is not desired.  This is ideal for optimizing back-office work; %They complete a data processing task in background. They are ideal for reducing work of back-office employees;
\item \textit{Hybrid}: Combination of the two above categories. %to provide automation for both front office and back office activities, allowing end-to-end automation of a process, through a human-robot interaction.
\end{itemize}
\item \textbf{Routine composition}: The ability of the RPA tool to orchestrate, at run-time, through \emph{manual support} or in an \emph{automated way} different (single) routines associated with different SW robots, when large workflows need to be emulated.

%
%\begin{itemize}
%\item \textit{Manual-based}: the most of RPA tools offer a scheduler that allows the coordination of a set of routines in order to emulate a workflow;
%\item \textit{Automated-based}: instead of delegating the orchestration to human supervisors, the management and the adaptation of complex workflows can be done in a dynamic way at run time, during the run of the system.%We tackled this problem in Section \ref{discussion}.
%\end{itemize}
\item \textbf{Log quality}: The quality of the logs recorded by the RPA tool (called \emph{RPA logs}). Since routines consist of collections of activities to be enacted according to certain routing constraints, logs produced by RPA tools resemble \emph{event logs} in process mining. To this end, we measure the quality of such logs using the classification provided in the Process Mining Manifesto \cite{van2011process}, where five maturity levels are defined, ranging from logs of excellent ($ \bigstar \bigstar \bigstar \bigstar \bigstar$) to poor quality ($\bigstar$).
\end{itemize}

\noindent
Table \ref{rpa-table} shows the results of the application of our classification framework to the selected RPA tools. The following aspects become apparent: the majority of the tools provide \textit{(i)} a Client-Server SW architecture, \textit{(ii)} GUIs with drag \& drop facilities and low-code functionalities, \textit{(iii)} both web and desktop recording, \textit{(iv)} a hybrid automation type, \textit{(v)} manual-based features to achieve routine composition, \textit{(vi)} logs of poor quality.
Interestingly, differently from the other tools, G1ANT and TagUI offer strong-coding functionalities with a basic CLI to support the programming of  SW robots. Finally, there is no tool that provides self learning or automated routine composition features.

\section{Research Challenges and Approaches}
\label{challenges-and-approaches}

On the basis of the results discussed in the previous section, we have derived four research challenges (and potential approaches to tackle them) necessary to inject intelligence into the current RPA technology towards a better support to BPM. The four identified challenges, which will be explained in the rest of the section, are: \emph{(i)} Intra-routine Self Learning, \emph{(ii)} Inter-routine Self Learning, \emph{(iii)} Automated generation of flowcharts, and \emph{(iv)} Automated routine composition.

\begin{enumerate}
\item \textbf{Intra-routine Self Learning (Segmentation)}. 

    \noindent
    \textit{Description}: Logs recorded by RPA tools are characterized by long sequences of actions and/or events that reflect a number of routine executions. A log can record information about several routines, whose actions and events are mixed in some order that reflects the particular order of their execution by the user \cite{RPM}. In addition, the same routine can be spread across multiple logs, making the automated identification of routines far from being trivial.

    \noindent
    \textit{Objective}: Identify the routines to be (potentially) emulated through software robots by looking at the RPA logs that keep track of the user actions taking place during a run of the system. This issue is known as ``\emph{segmentation}''.
    
    \noindent
    \textit{Approaches}:
    One possible approach to tackle this challenge is to rely on log analysis solutions in the Human-Computer Interaction (HCI) field \cite{dumais2014understanding,Dev:2017,Marrella:2018}, which focus on identifying frequent user tasks inside logs consisting of actions at different granularity. Alternately, local process mining approaches \cite{tax2016mining} or sequential pattern mining \cite{Dong:2009}
    can be employed to identify sequential patterns of non-consecutive actions that tend to be repeated multiple times across multiple logs \cite{RPM}.  An interesting recent approach is the one of Gao et al. \cite{COOPIS2019}, where the authors present a learning-based approach that allows for completely automated RPA-rule deduction, on the basis of captured historical low-level user behavior.
    However, to date, no available solution exists that allows for automatically: \emph{(i)} understanding which user actions have to be considered inside the log (separating noise from actions that contribute to routines); \emph{(ii)} interpreting their semantics on the basis of their granularity and \emph{(iii)} identifying which routines they belong to.
    Solving the above challenges would allow us to cluster all user actions associated with a routine in a well bounded execution trace. Consequently, all such execution traces would be organized into a \emph{routine-based log}.
\item \textbf{Inter-routine Self Learning (Automated identification of candidate routines to robotize)}. 

    \noindent
    \textit{Description}: While existing RPA tools allow one to automate a wide range of routines, they do not allow one to determine in an automated way which routines are good candidates for automation in the first place.

    \noindent
    \textit{Objective}: Given a list of routine-based logs, identify automatically which routines are good candidates for being automated by RPA tools.

    \noindent
    \textit{Approaches}: To date, current RPA tools provide very limited support to this challenge, which is often performed by means of interviews, walkthroughs, direct observation of workers, and analysis of documentation that may be of poor quality and difficult to understand. This manual approach allows analysts to identify the most obvious routines, while it is not suitable to detect those routines that are not executed on a daily basis or that are performed across multiple business units in different ways. The work of Jimenez-Ramirez and Reijers \cite{10.1007/978-3-030-21290-2_28} proposes to mitigate this issue through a method to improve the early stages of the RPA lifecycle using process mining techniques \cite{Aalst16}. On the other hand, a potential concrete solution to tackle this challenge is proposed by Bosco et al. \cite{RPM}, where the authors present a method to analyze routine-based logs in order to discover routines that are fully deterministic. To this end, the method combines a technique for compressing a set of routines into an acyclic automaton, with techniques for rule mining and for discovering data transformations.
\item \textbf{Automated generation of flowcharts.} 

    \noindent
    \textit{Description}: In RPA tools, there is a lacking of testing environments. As a consequence, SW robots are developed through a \emph{trial-and-error} approach consisting of three steps that are repeated until success \cite{mariaMaggi2018}: \emph{(i)} First, a human designer produces a flowchart diagram that includes the actions to be performed by the SW robot on a target system; \emph{(ii)} Second, SW robots are typically deployed in production environments, where they interact with information systems, with a high risk of errors due to inaccurate modeling of flowcharts; \emph{(iii)} Third, if SW robots are not able to reproduce the behavior of the users for a specific routine, then the designer adjusts the flowchart diagrams to fix the identified gap.
    While this approach is effective to execute simple rule-based logic in situations where there is no room for interpretation, it becomes time-consuming and error-prone in the presence of routines that are less predictable or require some level of human judgement. Indeed, the designer should have a global vision of all possible unfoldings of the routines to define the appropriate behaviors of the SW robot, which becomes complicated when the number of unfoldings increases. In cases where the rule set does not contain a suitable response for a specific situation, robots allow for escalation to a human supervisor.

    \noindent
    \textit{Objective}: Once the routines to be automated and the user actions that constitute them (i.e., the routine-based logs) have been identified, the target is to automatically generate the flowchart diagrams describing the behaviors of the SW robots required to successfully execute the routines.

\begin{figure*}[t!]
\centering
\caption{Overview of the pipeline of potential approaches required to tackle the research challenges}
\includegraphics[width=\textwidth,keepaspectratio=true]{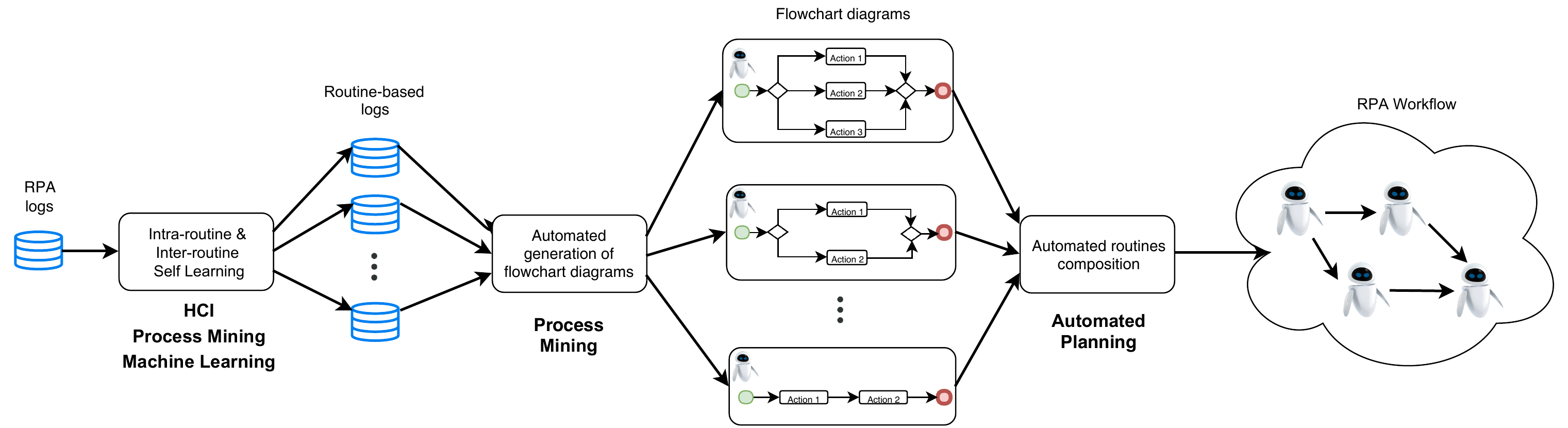}
\label{fig:approach_overview}
\end{figure*}

    \noindent
    \textit{Approaches}: 
    A possible solution to this challenge can be to resort on discovery algorithms from the process mining field \cite{Aalst16} and to automatically extract flowcharts in the form of Petri nets/BPMN models from routine-based logs. Thus, it is necessary to investigate from the literature on \emph{process discovery} \cite{DBLP:journals/corr/AugustoCDRMMMS17} which algorithms suit better to extract the base structure of flowchart diagrams from a routine-based log. Since such discovered flowcharts will reflect real routine executions, they may contain/miss (un-)necessary user actions with respect to the SW robot's expected behavior. To mitigate this issue, it is possible to leverage another process mining technique, named \emph{trace alignment} \cite{Adriansyah2011CostBasedFI}, which would allow us to properly clean the discovered flowchart diagrams, by filtering out the unwanted behaviors found in the previous discovery stage.
\item \textbf{Automated routines composition}.

    \noindent
    \textit{Description}: In modern contexts, human operators usually enact not just single tasks but complex workflows, consisting of many interrelated routines. However, the current RPA technology allows for developing SW robots for executing single, independent routines. Only manual support is provided to orchestrate multiple routines, i.e., the management of more complex workflows is completely delegated to human supervisors. 
    
    \noindent
    \textit{Objective}: Automated generation of RPA workflows consisting of many interrelated routines.
    
    \noindent
    \textit{Approaches}:
    To synthesize complex workflows through an intelligent orchestration of the robots' routines, \emph{automated planning} techniques in AI can be employed \cite{ghallab2004automated}. The application of planning for tackling the composition issue has been already proved to be successful in real world domains \cite{Marrella19}. The idea is to consider the robots' routines as black boxes, i.e., as planning actions with specific preconditions and effects, and to delegate to a planning system the generation of a proper strategy to automatically compose them in a larger workflow that coordinates their orchestration.    
\end{enumerate} 

Figure \ref{fig:approach_overview} shows a graphical overview of the pipeline of potential approaches required to tackle the four identified research challenges to achieve the aforementioned objectives.

\section{Discussion and Concluding Remarks}
\label{sec:conclusion}

RPA recently gained a lot of attention in the BPM domain \cite{vanderAalst2018}. Since RPA operates at the UI level, rather than at the system level, it allows one to apply automation without any changes in the underlying information system. Thus, the entry barrier of adopting RPA in BPs that are already in place is lower compared to conventional BPM \cite{COOPIS2019}. However, the current generation of RPA tools is driven by predefined rules and manual configurations made by expert users rather than by AI \cite{aitiptoes}, preventing a widespread adoption of these tools in the BPM domain.

In this paper, we have tackled this issue starting from an in-depth experimentation of the RPA tools available in the market. Then we have provided a classification framework to categorize them on the basis of some key dimensions and we have derived four research challenges and discussed potential approaches necessary to inject intelligence into the current RPA technology, from a BPM perspective.

It is worth to notice that, according to Table \ref{rpa-table}, the logs produced by the tested RPA tools have a poor quality (actions may be missing or not recorded properly), since they are mainly used for debugging purposes. Increasing the quality of RPA logs is a fundamental prerequisite to properly tackle the proposed research, which leverages a log analysis to discover, identify, model and compose routines in an automated
way. To this end, RPA tools should aim at logs at the highest possible quality level. 

To mitigate this issue, we are currently developing an action logger to be attached to the existing RPA tools, in order to enable the creation of routines-based logs of an acceptable quality. Apart from the need to increase the quality of RPA logs, as a future work, this research aims at also improving the \emph{auditability} (RPA logs are auditable), \emph{upgradability} (flowchart diagrams describing SW robots' behavior will be always updated to the current state of the system execution) and the \emph{resiliency} (SW robots will be always upgraded to deal with new behaviors, making them very robust to any contextual change that may arise during a routine execution) of
SW robots. Furthermore, \emph{scalability} must be improved as well. Human capacity is difficult to scale in
situations where demand fluctuates, instead SW robots operate at whatever speed is demanded by
the work volume.

To conclude, we note that our study has a threat to validity, since we analyzed only a sample of the RPA tools available on the market. As a consequence, our findings can not be generalized beyond the scope of the tested RPA tools. Nonetheless, we consider this work as an important first step towards the realization of intelligent solutions for RPA.
Moreover, we also envision that this research will provide long-term benefits on the
companies workforce, e.g., by improving the customer service in the front office while at the same time
reducing the back office tasks. 

\bibliographystyle{aaai}
\bibliography{main}

\begin{thebibliography}{}

\bibitem[\protect\citeauthoryear{Adriansyah, Sidorova, and van
  Dongen}{2011}]{Adriansyah2011CostBasedFI}
Adriansyah, A.; Sidorova, N.; and van Dongen, B.~F.
\newblock 2011.
\newblock {Cost-Based Fitness in Conformance Checking}.
\newblock {\em 2011 Eleventh International Conference on Application of
  Concurrency to System Design}  57--66.

\bibitem[\protect\citeauthoryear{Aguirre and Rodriguez}{2017}]{Aguirre2017}
Aguirre, S., and Rodriguez, A.
\newblock 2017.
\newblock {Automation of a Business Process Using Robotic Process Automation
  (RPA): A Case Study}.
\newblock In {\em Applied Computer Sciences in Engineering},  65--71.
\newblock Cham: Springer International Publishing.

\bibitem[\protect\citeauthoryear{AI-Multiple}{2019}]{RPA}
AI-Multiple.
\newblock 2019.
\newblock {All 52 RPA Software Tools and Vendors: Sortable List [2019]}.
\newblock \url{https://blog.aimultiple.com/rpa-tools/}.

\bibitem[\protect\citeauthoryear{Augusto \bgroup et al\mbox.\egroup
  }{2019}]{DBLP:journals/corr/AugustoCDRMMMS17}
Augusto, A.; Conforti, R.; Dumas, M.; Rosa, M.~L.; Maggi, F.~M.; Marrella, A.;
  Mecella, M.; and Soo, A.
\newblock 2019.
\newblock {Automated Discovery of Process Models from Event Logs: Review and
  Benchmark}.
\newblock {\em TKDE} 31(4).

\bibitem[\protect\citeauthoryear{Bisbal \bgroup et al\mbox.\egroup
  }{1999}]{bisbal1999legacy}
Bisbal, J.; Lawless, D.; Wu, B.; and Grimson, J.
\newblock 1999.
\newblock {Legacy information systems: Issues and directions}.
\newblock {\em IEEE Software} 16(5):103--111.

\bibitem[\protect\citeauthoryear{Bosco \bgroup et al\mbox.\egroup }{2019}]{RPM}
Bosco, A.; Augusto, A.; Dumas, M.; La~Rosa, M.; and Fortino, G.
\newblock 2019.
\newblock {Discovering Automatable Routines From User Interaction Logs}.
\newblock In {\em 17th International Conference on Business Process Management
  (BPM'19), Forum track, Vienna, Austria},  144--162.
\newblock Cham: Springer International Publishing.

\bibitem[\protect\citeauthoryear{Dev and Liu}{2017}]{Dev:2017}
Dev, H., and Liu, Z.
\newblock 2017.
\newblock {Identifying Frequent User Tasks from Application Logs}.
\newblock In {\em Proceedings of the 22Nd International Conference on
  Intelligent User Interfaces}, IUI '17,  263--273.
\newblock New York, NY, USA: ACM.

\bibitem[\protect\citeauthoryear{Dong}{2009}]{Dong:2009}
Dong, G.
\newblock 2009.
\newblock {\em {Sequence Data Mining}}.
\newblock Berlin, Heidelberg: Springer-Verlag.

\bibitem[\protect\citeauthoryear{Dumais \bgroup et al\mbox.\egroup
  }{2014}]{dumais2014understanding}
Dumais, S.; Jeffries, R.; Russell, D.~M.; Tang, D.; and Teevan, J.
\newblock 2014.
\newblock {Understanding User Behavior Through Log Data and Analysis}.
\newblock In {\em Ways of Knowing in HCI}. New York, NY: Springer.
\newblock  349--372.

\bibitem[\protect\citeauthoryear{Gao \bgroup et al\mbox.\egroup
  }{2019}]{COOPIS2019}
Gao, J.; van Zelst, S.~J.; Lu, X.; and van~der Aalst, W. M.~P.
\newblock 2019.
\newblock Automated robotic process automation: A self-learning approach.
\newblock In {\em On the Move to Meaningful Internet Systems: OTM 2019
  Conferences},  95--112.
\newblock Cham: Springer International Publishing.

\bibitem[\protect\citeauthoryear{Geyer-Klingeberg \bgroup et al\mbox.\egroup
  }{2018}]{geyer2018process}
Geyer-Klingeberg, J.; Nakladal, J.; Baldauf, F.; Veit, F.; van~der Aalst, W.;
  Casati, F.; Conforti, R.; de~Leoni, M.; and Dumas, M.
\newblock 2018.
\newblock {Process Mining and Robotic Process Automation: A Perfect Match}.
\newblock In {\em 16th International Conference on Business Process Management
  (BPM'18), Dissertation/Demos/Industry track, Sidney, Australia},  124--131.

\bibitem[\protect\citeauthoryear{Ghallab, Nau, and
  Traverso}{2004}]{ghallab2004automated}
Ghallab, M.; Nau, D.; and Traverso, P.
\newblock 2004.
\newblock {\em Automated Planning: theory and practice}.
\newblock Elsevier.

\bibitem[\protect\citeauthoryear{Hill, Ford, and Farreras}{2015}]{hill2015real}
Hill, J.; Ford, W.~R.; and Farreras, I.~G.
\newblock 2015.
\newblock Real conversations with artificial intelligence: A comparison between
  human--human online conversations and human--chatbot conversations.
\newblock {\em Computers in Human Behavior} 49:245--250.

\bibitem[\protect\citeauthoryear{Jimenez-Ramirez \bgroup et al\mbox.\egroup
  }{2019}]{10.1007/978-3-030-21290-2_28}
Jimenez-Ramirez, A.; Reijers, H.~A.; Barba, I.; and Del~Valle, C.
\newblock 2019.
\newblock {A Method to Improve the Early Stages of the Robotic Process
  Automation Lifecycle}.
\newblock In {\em 31st International Conference on Advanced Information Systems
  Engineering (CAiSE'19), Rome, Italy},  446--461.
\newblock Cham: Springer International Publishing.

\bibitem[\protect\citeauthoryear{Kirchmer}{2017}]{kirchmer2017robotic}
Kirchmer, M.
\newblock 2017.
\newblock {Robotic Process Automation-Pragmatic Solution or Dangerous
  Illusion}.
\newblock {\em BTOES Insights, June'17}.

\bibitem[\protect\citeauthoryear{Lacity, Willcocks, and
  Craig}{2015}]{lacity2015robotic}
Lacity, M.; Willcocks, L.~P.; and Craig, A.
\newblock 2015.
\newblock {\em {RPA at Telefonica O2}}.
\newblock The London School of Economics and Political Science.

\bibitem[\protect\citeauthoryear{Leno \bgroup et al\mbox.\egroup
  }{2018}]{mariaMaggi2018}
Leno, V.; Dumas, M.; Maggi, F.~M.; and La~Rosa, M.
\newblock 2018.
\newblock Multi-perspective process model discovery for robotic process
  automation.
\newblock {\em CAiSE Doct. Cons.}

\bibitem[\protect\citeauthoryear{Leno \bgroup et al\mbox.\egroup
  }{2019}]{lenoaction}
Leno, V.; Polyvyanyy, A.; Rosa, M.~L.; Dumas, M.; and Maggi, F.~M.
\newblock 2019.
\newblock Action logger: Enabling process mining for robotic process
  automation.
\newblock In {\em Proceedings of the Dissertation Award, Doctoral Consortium,
  and Demonstration Track at 17th International Conference on Business Process
  Management, ({BPM}'19), Vienna, Austria},  124--128.

\bibitem[\protect\citeauthoryear{Lohr}{2018}]{aitiptoes}
Lohr, S.
\newblock 2018.
\newblock {The Beginning of a Wave: A.I. Tiptoes Into the Workplace}.
\newblock
  \url{https://www.nytimes.com/2018/08/05/technology/workplace-ai.html/}.

\bibitem[\protect\citeauthoryear{Marrella and Catarci}{2018}]{Marrella:2018}
Marrella, A., and Catarci, T.
\newblock 2018.
\newblock {Measuring the Learnability of Interactive Systems Using a Petri Net
  Based Approach}.
\newblock In {\em Proceedings of the 2018 Designing Interactive Systems
  Conference}, DIS '18,  1309--1319.
\newblock New York, NY, USA: ACM.

\bibitem[\protect\citeauthoryear{Marrella}{2019}]{Marrella19}
Marrella, A.
\newblock 2019.
\newblock {Automated Planning for Business Process Management}.
\newblock {\em J. Data Semantics} 8(2):79--98.

\bibitem[\protect\citeauthoryear{Reichert and Weber}{2012}]{ReichertBook2012}
Reichert, M., and Weber, B.
\newblock 2012.
\newblock {\em {Enabling Flexibility in Process-Aware Information Systems -
  Challenges, Methods, Technologies}}.
\newblock Springer Berlin Heidelberg.

\bibitem[\protect\citeauthoryear{SE}{2019}]{celonis}
SE, C.
\newblock 2019.
\newblock {Academic Alliance}.
\newblock \url{https://www.celonis.com/academic-alliance}.

\bibitem[\protect\citeauthoryear{Tax \bgroup et al\mbox.\egroup
  }{2016}]{tax2016mining}
Tax, N.; Sidorova, N.; Haakma, R.; and van~der Aalst, W.~M.
\newblock 2016.
\newblock Mining local process models.
\newblock {\em Journal of Innovation in Digital Ecosystems} 3(2):183--196.

\bibitem[\protect\citeauthoryear{Tornbohm}{2017}]{market}
Tornbohm, C.
\newblock 2017.
\newblock {Gartner market guide for Robotic Process Automation software}.
\newblock Report G00319864. Gartner.

\bibitem[\protect\citeauthoryear{van~der Aalst and van
  Hee}{2004}]{van2004workflow}
van~der Aalst, W., and van Hee, K.
\newblock 2004.
\newblock {\em Workflow management: models, methods, and systems}.
\newblock MIT press.

\bibitem[\protect\citeauthoryear{van~der Aalst \bgroup et al\mbox.\egroup
  }{2012}]{van2011process}
van~der Aalst, W.; Adriansyah, A.; de~Medeiros, A. K.~A.; Arcieri, F.; Baier,
  T.; Blickle, T.; Bose, J.~C.; van~den Brand, P.; Brandtjen, R.; Buijs, J.;
  Burattin, A.; Carmona, J.; Castellanos, M.; Claes, J.; Cook, J.; Costantini,
  N.; Curbera, F.; Damiani, E.; de~Leoni, M.; Delias, P.; van Dongen, B.~F.;
  Dumas, M.; Dustdar, S.; Fahland, D.; Ferreira, D.~R.; Gaaloul, W.; van
  Geffen, F.; Goel, S.; G{\"u}nther, C.; Guzzo, A.; Harmon, P.; ter Hofstede,
  A.; Hoogland, J.; Ingvaldsen, J.~E.; Kato, K.; Kuhn, R.; Kumar, A.; La~Rosa,
  M.; Maggi, F.; Malerba, D.; Mans, R.~S.; Manuel, A.; McCreesh, M.; Mello, P.;
  Mendling, J.; Montali, M.; Motahari-Nezhad, H.~R.; zur Muehlen, M.;
  Munoz-Gama, J.; Pontieri, L.; Ribeiro, J.; Rozinat, A.; Seguel~P{\'e}rez, H.;
  Seguel~P{\'e}rez, R.; Sep{\'u}lveda, M.; Sinur, J.; Soffer, P.; Song, M.;
  Sperduti, A.; Stilo, G.; Stoel, C.; Swenson, K.; Talamo, M.; Tan, W.; Turner,
  C.; Vanthienen, J.; Varvaressos, G.; Verbeek, E.; Verdonk, M.; Vigo, R.;
  Wang, J.; Weber, B.; Weidlich, M.; Weijters, T.; Wen, L.; Westergaard, M.;
  and Wynn, M.
\newblock 2012.
\newblock {Process Mining Manifesto}.
\newblock In {\em Business Process Management Workshops},  169--194.
\newblock Berlin, Heidelberg: Springer Berlin Heidelberg.

\bibitem[\protect\citeauthoryear{van~der Aalst, Bichler, and
  Heinzl}{2018}]{vanderAalst2018}
van~der Aalst, W. M.~P.; Bichler, M.; and Heinzl, A.
\newblock 2018.
\newblock {Robotic Process Automation}.
\newblock {\em Business {\&} Information Systems Engineering} 60(4):269--272.

\bibitem[\protect\citeauthoryear{van~der Aalst}{2016}]{Aalst16}
van~der Aalst, W. M.~P.
\newblock 2016.
\newblock {\em {Process Mining: Data Science in Action}}.
\newblock Heidelberg: Springer, 2 edition.

\bibitem[\protect\citeauthoryear{Willcocks, Lacity, and
  Craig}{2015}]{willcocks2015function}
Willcocks, L.~P.; Lacity, M.; and Craig, A.
\newblock 2015.
\newblock {\em {The IT Function and Robotic Process Automation}}.
\newblock The London School of Economics and Political Science.

\bibitem[\protect\citeauthoryear{Willcocks}{2016}]{willcocks2016service}
Willcocks, L.
\newblock 2016.
\newblock {\em {Service Automation : robots and the future of work}}.
\newblock Warwickshire, United Kingdom: Steve Brookes Publishing.

\end{thebibliography}

\end{document}